\documentclass{article}
\usepackage{spconf,amsmath,graphicx,amssymb}
\usepackage{url}

\usepackage{enumitem}
\usepackage{pbox}
\setlist{nosep, leftmargin=14pt}

\usepackage{mwe} 
\usepackage{xurl}

\usepackage{hyperref}

\title{Weakly supervised localisation of prostate cancer using reinforcement learning for bi-parametric MR images}

\name{\begin{tabular}{@{}c@{}}
{Martynas Pocius{${^{1}}$} } \qquad
{Wen Yan{${^{1}}$}} \qquad
{Dean C. Barratt{${^{1}}$}} \qquad
{Mark Emberton{${^{1}}$} } \qquad\\
\textit{{Matthew J. Clarkson{${^{1}}$}} \qquad
{Yipeng Hu{${^{1}}$}} \qquad
{Shaheer U. Saeed{${^{1}}$}}}
\end{tabular}
}

\address{
    {${^{1}}$} Centre for Medical Image Computing and Wellcome/EPSRC Centre for Interventional \\
    and Surgical Sciences, University College London, London, UK \\
    {${^{2}}$} Department of Urology, University College London Hospitals NHS Foundation Trust, London, UK
}

\begin{document}

\maketitle

\begin{abstract}
    In this paper we propose a reinforcement learning based weakly supervised system for localisation. We train a controller function to localise regions of interest within an image by introducing a novel reward definition that utilises non-binarised classification probability, generated by a pre-trained binary classifier which classifies object presence in images or image crops. The object-presence classifier may then inform the controller of its localisation quality by quantifying the likelihood of the image containing an object. Such an approach allows us to minimize any potential labelling or human bias propagated via human labelling for fully supervised localisation. We evaluate our proposed approach for a task of cancerous lesion localisation on a large dataset of real clinical bi-parametric MR images of the prostate. Comparisons to the commonly used multiple-instance learning weakly supervised localisation and to a fully supervised baseline show that our proposed method outperforms the multi-instance learning and performs comparably to fully-supervised learning, using only image-level classification labels for training. 
\end{abstract}


\section{Introduction}
\label{sec:intro}

Recently medical imaging tasks are increasingly being automated using supervised deep learning, which utilises expert annotated data for training \cite{yoo2019prostate,goldenberg2019a}. One such task is prostate cancer localisation using bi-parametric MR images, of interest in this work. This image-based cancer localisation has recently received attention since it provides an alternative to invasive biopsies for diagnosing cancer. The task, however, may prove challenging even when expert labels are available, as indicated by the reported 7-14\% missed clinically significant cancers in routine practice \cite{ahmed2017diagnostic,rouviere2019use} and by large disagreement between experts \cite{chen2020variability}. Low dice scores in the range of 20-40\% are common for recently reported works that automate the task of cancerous lesion localisation \cite{saeed2022image_overlapping,yan2022impact}. This may be caused by the task itself being challenging with challenges further exacerbated by label quality issues e.g.,inter- and intra-observer variance and potential bias due to inconsistent reporting guidelines \cite{wang2019deeply}.

In contrast to fully supervised learning, weak supervision may allow for localisation of features without any human labels of localisation \cite{otalora2021combining}. Such paradigms alleviate the need for expensive data annotation with regards to localisation, which may be replaced with annotation with respect to object presence. Not only is human annotation for localisation expensive, it may introduce biases in the form of inter-observer variances due to the difficulty in pixel-level annotating \cite{shin2019joint,yiqiu2021interpretable}. In contrast to fully supervised approaches which operate in an end-to-end black box manner, weakly supervised approaches may offer intermediate signals, such as the likelihood of object presence within the localised area aiding explainability.

A weak supervision system works by using weak labels of object presence (binary image-level labels) to generate the likelihood of a region containing an object - which may be referred to as the `object-ness' of the region \cite{yiqiu2021interpretable,shin2019joint,ren2016weakly}. The object-ness score may be generated by means of the non-binarised class probabilities of an object presence classification function, such as a neural network. This then informs the localisation by quantifying the object-ness of each region and selecting, by means of controller function, the region with the highest object-ness. This region may then be considered to contain the object of interest.

Even though weak supervision systems solve some issues in supervised learning, there are certain cases when conventional weak supervision models cannot be used - some objectives do not have clearly definable loss functions, let alone making them differentiable. In our work, we use reinforcement learning (RL) as weakly supervised way of a training a localisation system. In this system, the controller learns localisation policy through a reward based on object-ness which is learnt via classification labels of object presence. Such an indirect training approach solves issues arising in cases of non-differentiable objectives while not utilising any localisation labels for training, which may introduce human biases \cite{czolbe2021segmentation}. 

Although RL has not been widely explored in weak supervision disease localisation on medical images, there have been a number of attempts. Ali et al. \cite{ali2018lung} proposed RL for lung nodule detection using CT. The final system outperformed compared baseline systems. In another, study Pesce et al. \cite{emanuele2019learning} proposed a REINFORCE-algorithm-based system for pulmonary lesion detection in chest radiographs. The final system achieved 45\% average overlap outperforming baseline models. Nevertheless, most studies base training on overlap-based measures which may introduce additional supervision and the requirement for human labels of localisation.

In this study, we propose a RL-based weakly supervised system for object localisation. The localisation-performing controller is trained using a reward based on object-ness (similar to previous task-based rewards \cite{saeed2021adaptable,yoon2019data}). The object-ness is based on non-binarised classification probabilities generated by a pre-trained classifier, which was trained to predict object presence using only image level binary labels of presence. Non-binarised classification probability of this classifier serves as a proxy for the object-ness for images or regions/ crops of images. The controller may localise a region of an image. For the localised region, the object-ness-predicting classifier computes the object-ness which is formulated into a reward to train the controller. Such an approach produces an effective localisation policy that can identify regions of interest, without using any human labels of localisation. We evaluate our proposed approach for a task of malignant cancerous lesion localisation for bi-parametric MR images. The only labels required to train a localisation system for lesions (defined as predicting bounding boxes) are the labels of lesion presence at the 2D bi-parametric MR image level.

\textbf{Contributions:} 1) we present a novel reward mechanism to train a RL-based weakly supervised localisation using an object-ness predicting classifier; 2) we use a clinically challenging task of prostate cancer localisation using a large dataset of real clinical bi-parametric MR for evaluation, utilising only image-level labels of cancer presence for training; 3) we make our implementation publicly available: \url{https://github.com/MartynasPocius/PPO-Localising-Prostate-Cancer}.

\section{Methods}
\label{sec:format}

\subsection{Object-ness predicting classifier}
\label{ssec:classifier}

The object-ness predicting classifier $f(\cdot; w): \mathcal{X} \rightarrow [0, 1]$ outputs an object-ness score for a given sample $x\in\mathcal{X}$, and has weights $w$. Here, $\mathcal{X}$ denotes the domain of image samples. The optimal weights for the $f$ are obtained by training using a dataset of image-label pairs indicating object presence within a given image i.e., $\{x_i, y_i\}_{i=1}^N$, where $y_i\in\{0,1\}$ such that $y_i = 1$ if $x_i$ contains the object of interest and $y_i = 0$ otherwise. The non-binarised prediction probability of this classifier function serves as the object-ness score in this work.

The following binary cross-entropy loss function may be used to train the classifier such that non-binarised probabilities represent the object-ness of an image:

\begin{align}
    & l(y_i, f(x_i; w)) = \notag\\
    & -\frac{1}{N}\sum_{i=1}^N (y_i log(f(x_i; w))+(1-y_i)log(1-f(x_i; w)),
\end{align}

where $y_i$ is the binary label of object presence for image $x_i$ and $f(x_i; w)$ is a prediction from the classifier for the same sample. The optimisation problem to obtain optimal weights $w^*$ for classifier $f$ is:

\begin{equation}
    w^* = \mathop{\text{argmin}}_w \mathbb{E} [l(y, f(x; w))]
\end{equation}

After training, weights of $f$ may be fixed as $w^*$ and non-binarised prediction probabilities $f(x_i, w)$ may be used as a proxy for the object-ness of an image sample. 




\begin{figure}
    \centering
    \includegraphics[width=0.46\textwidth]{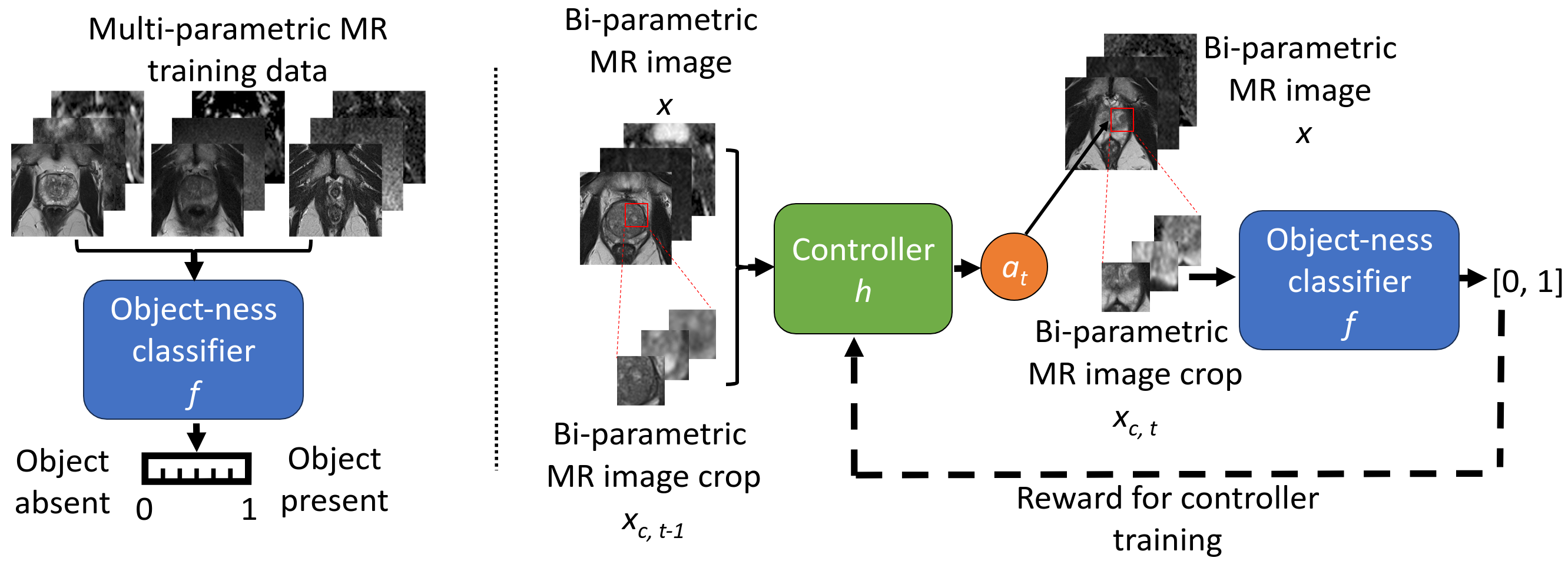}
    \caption{Overview of the proposed method.}
    \label{fig:methods}
\end{figure}

\subsection{The controller function for localisation}\label{sec:methods_controller}

We consider an image $x\in\mathcal{X}$ to be a set of pixel values, where a subset $x_c \subset x$ is considered a region or crop of the sample. When a crop is made with respect to cropping parameters $a$ (which may specify location and size of the crop), we denote the crop as $x_{c|{a_t}} \subset x$. It should be noted that as with $x_c\in\mathcal{X}_c$ where $\mathcal{X}_c$ may be considered a special case of $\mathcal{X}$.

The controller function $h(\cdot; \theta): \mathcal{X} \times \mathcal{X} \rightarrow \mathcal{A}$, has weights $\theta$, and outputs the cropping parameters $a_t\in\mathcal{A}\in\mathbb{R}^4$, given the image sample $x_t\in\mathcal{X}$ and the previously localised crop $x_{c|{a_{t-1}}, ~t-1}\in\mathcal{X}$, for time-step $t$. The action $a$ is 4-dimensional since it consists of the location at which to crop, in the horizontal-vertical coordinate space as well as the horizontal and vertical size for the crop.

\subsection{The controller-environment interactions}\label{sec:methods_env}

The controller function is trained using RL, where the RL environment consists of the image and the object-ness predicting classifier. The observed state at time-step $t$ is defined as $s_t = \{x_t, x_{c|{a_{t-1}}, ~t-1}\}$, and the action is defined as $a_t\in\mathcal{A}$ (which consists of the location and size of cropping), where $\mathcal{A}$ is the action space. For notational convenience in further analysis we denote the domain from which states $s_t$ are sampled, or the state space, as $s\in\mathcal{S}$, therefore, $\mathcal{X}\times \mathcal{X} \equiv \mathcal{S}$. 

The environment may be defined as a Markov decision process and denoted as $(\mathcal{S}, \mathcal{A}, p, r, \pi, \gamma)$. The state transition distribution conditioned on state-action pairs is given by $p:\mathcal{S} \times \mathcal{S} \times \mathcal{A} \rightarrow [0, 1]$, where $p(s_{t+1} | s_t, a_t)$ denotes the probability of the next state $s_{t+1}$, given the current state $s_t$ and action $a_t$.

In our formulation, the reward function $r:\mathcal{S}\times\mathcal{A}\rightarrow \mathbb{R}$ produces a reward $R_t = r(s_t, a_t)$ at time-step $t$. The reward is computed using the object-ness predicting classifier $f(x_{c|{a_{t}}, ~t}; w^*)$, where the image $x_t$ is cropped, based on cropping parameters $a_t$ and passed to the pre-trained and fixed object-ness predicting classifier $f$, with optimal parameters $w^*$. Therefore, we note that $f(x_{c|{a_{t}}, ~t}; w^*) \equiv r(s_t, a_t)$, where $s_t = \{x_t, x_{c|{a_{t-1}}, ~t-1}\}$, as defined above.

Policy $\pi:\mathcal{S}\times\mathcal{A}\rightarrow [0, 1]$ denotes probability of performing an action $a_t$ given the state $s_t$ i.e., $\pi(a_t|s_t)$ for time-step $t$. The action is sampled according to the policy $a_t \sim \pi(\cdot)$.

Following the policy for sampling actions and the state transition distribution for sampling next states and the reward function, we can generate a trajectory of states, actions and rewards over multiple time-steps $(s_1, a_1, R_1, \dots, s_T, a_T, R_T)$.

We consider the policy to be modelled as a neural network with parameters $\theta$, i.e., $\pi(\cdot; \theta)$. In practice, the neural network may predict parameters of a distribution e.g., a Gaussian distribution from which to sample actions \cite{schulman2017proximal}. The goal using RL is to find optimal policy parameters $\theta^*$, such that following the policy $\pi(\cdot; \theta^*)$ maximises the accumulated reward.

In order to find the optimal policy parameters $\theta^*$, we compute the cumulative reward over the trajectory over multiple time-steps: $Q^{\pi(\cdot; \theta)}(s_t, a_t) = \sum^T_{k=0} \gamma^k R_{t+k}$, where $\gamma$ is the discount factor used to weight future rewards. The central optimisation problem may be solved using gradient ascent:

\begin{equation}
    \theta^* = \mathop{\text{argmax}}_\theta \mathbb{E} [Q^{\pi(\cdot; \theta)}(s_t, a_t)]
\end{equation}


\subsection{Using the trained controller for localisation}

Once the controller is trained, and optimal parameters $\theta^*$ are found, the controller may be used to sample actions that effectively localise the objects of interest within images by following $a_t = h(s_t; \theta^*)$, where, we repeat this over $T$ time-steps at inference to generate our final localised region ($T=1024$ in our work). We only use one trajectory at inference to localise the region without averaging, common in MIL or some crop-based methods.

\section{Experiments}


\subsection{Data acquisition}

Bi-parametric 2D MR images (diffusion weighted imaging (DWI) with high b-values b$\geqslant$1400, apparent diffusion coefficient (ADC) and T2 weighted (T2W)) were acquired for 850 prostate cancer patients, as part of the clinical trials carried out at University College London Hospitals: SmartTarget \cite{hamid2019smarttarget}, PICTURE \cite{simmons2018accuracy}, ProRAFT \cite{orczyk2021prostate}, Index \cite{dickinson2013multi}, PROMIS \cite{bosaily2015promis} and PROGENY \cite{linch2017intratumoural}. All trial patients gave written consents and the ethics was approved as part of the respective trial protocols. After centre-cropping around the prostate gland, all image intensities were scaled to the range [0, 1]. To reduce task complexity, we only focused on one lesion at the time, therefore, slices containing multiple lesions have been removed. The final dataset consisted of $5978$ 2D $200\times200\times3$ slices (where the three channels correspond to three MR modalities), with 3586, 1196 and 1196 slices from 378, 126 and 126 patients (630 patients in total) in the train, validation and holdout sets.

\subsection{Network architectures and training details}

\subsubsection{Object-ness predicting classifier}

In this work, for the object-ness classifier $f$, we used the LeNet architecture \cite{lecun1998gradient}. $5$ convolutional blocks - 2D convolutional layer with $(3 \times 3)$ kernel and 2D Max Pooling with $(2 \times 2)$ kernel, were followed by a dense layer which led to the output layer that produces an output of [0, 1].

A mini batch size of 16 was used for training for 8 epochs, over an approximate time of 3 hours on a single Nvidia Tesla V100 GPU. The trained model achieved 99.7\% accuracy at the binary object presence classification task, for the task of cancer presence classification on image slices.

\subsubsection{Controller}

The controller followed the a CNN architecture similar to the object-ness classifier with 4 convolutional blocks and 2 dense layers. The proximal policy optimisation (PPO) \cite{schulman2017proximal} algorithm was used for training the controller, where default values for learning rate, batch size, and entropy coefficient ($0.0003$, $128$, $0.001$, respectively) were used. Changing these values did not alter performance, showing the method to be robust to changes in these hyper-parameters.

\subsection{Ablation studies}

During RL training, 3 different state-action configurations were tested (see Table ~\ref{tab2}). In configurations 1-2, state is defined as cropped image patch, while state in configuration 3 cropped image is concatenated with the original image.

In Configuration 1, the state is only the previous cropped patch $s_t = x_{c|{a_{t-1}}, ~t-1}$ and the action is the absolute location and size for the cropping parameters. In Configuration 2, the state is only the previous cropped patch $s_t = x_{c|_{a_{t-1}}, ~t-1}$ and the action is the change in location and size for the cropping parameters, compared to the previous time-step. In Configuration 3, the state is the concatenation of the un-cropped image with the obtained crop from the previous time-step $s_t = \{x_t, x_{c|{a_{t-1}}, ~t-1}\}$ and the action is the absolute location and size for the cropping parameters.

\subsection{Comparing with baselines}


\subsubsection{Fully Supervised}

This network followed the same architecture as the object-ness classifier. The output layer was changed to 4-dimensional to accommodate $a_t\in\mathcal{A}\in\mathbb{R}^4$ for the cropping parameters that specify horizontal-vertical crop location and size in two axes. The mean squared error (MSE) loss was used to train this network to localise cancerous regions.

\subsubsection{Multiple-Instance-Learning}
The weakly supervised MIL model \cite{patil2019breast,ren2016weakly} training was conducted with 'bags' (set of randomly cropped patches) - if any patches within the bag contain part of our area of interest, it is marked as positive \cite{ren2016weakly}. By learning bag classification, using a binary cross-entropy using bag-level labels of presence, the MIL model will learn the relationship between multiple patch features and bag labels \cite{ren2016weakly,patil2019breast}. For this study, each image array of shape $200\times200\times3$ was considered to be a bag containing $400$ randomly cropped $16\times16 \times 3$ patches. We follow a training strategy similar to Ren et al. \cite{ren2016weakly}. Although our proposed RL model does not use any localisation labels in training, the MIL system uses localisation labels for bag construction, to train the MIL bag-level classifier, to determine if any patch within the bag contains the ROI. However, we still argue that this is the most comparable previously proposed method to our proposed solution. Previously proposed methods all utilise localisation labels at some point during training to an even greater extent \cite{patil2019breast,stoyanov2018deep,ren2016weakly}, which may lead to an unfair comparison, where our proposed method does not use localisation labels at any point in the training.

\section{Results}
{
\setlength\intextsep{0pt}
\begin{table}[htb]
\caption{Performance on the holdout set.}\label{tab2}
  \resizebox{\columnwidth}{!}{
    \begin{tabular}{|c|c|c|c|}
    \hline
    Model & State & Action & Dice score\\
    \hline
    RL - Configuration 1 & previous crop & absolute & $0.31$ $\pm$ $0.08$\\
    RL - Configuration 2 & previous crop & relative change & $0.26$ $\pm$ $0.06$\\
    RL - Configuration 3 & previous crop + un-cropped image & absolute & $0.38$ $\pm$ $0.09$\\
    MIL & N/A & N/A & $0.25$ $\pm$ $0.11$\\
    Fully-supervised & N/A & N/A &  $0.34$ $\pm$ $0.07$\\
    \hline
    \end{tabular}
  }
\end{table}
}
\subsection{Ablation studies} 

The RL system achieved convergence before the 1000th iteration, in an average of 38 hours on a single Nvidia Tesla V100 GPU. Mean dice score is reported over the holdout set (summarised in Table~\ref{tab2}). Configuration 3 (with state consisting of both the un-cropped image the cropped patch from the previous time-step, and the action being an absolute value of the cropping parameters) achieved the highest dice score of $0.38 \pm 0.09$ compared with Configurations 1 and 2 which achieved $0.31 \pm 0.07$ and $0.26 \pm 0.06$ dice scores, respectively, with statistical significance (p-values $0.011$ and $0.007$ for Configurations 1 and 2 respectively).

\subsection{Baseline comparison}

Table~\ref{tab2} shows that our proposed method ($0.38 \pm 0.09$) outperformed MIL ($0.25 \pm 0.11$) with statistical significance (p-value$=0.008$). Although without the statistical significance (p-value$=0.092$), it demonstrated a better performance than the fully-supervised network ($0.34 \pm 0.07$).




\begin{figure}
    \centering
    \includegraphics[width=0.36\textwidth]{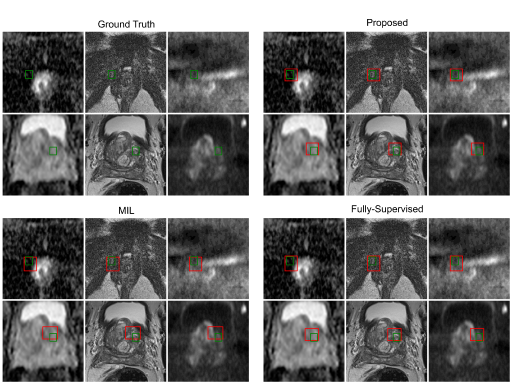}
    \caption{Samples (green: ground truth; red: predicted).}
    \label{fig:res}
\end{figure}

\section{Discussion and Conclusion}

Our proposed system learnt an effective localisation policy for prostate cancer localisation on bi-parametric MR, comparable to previous works in literature \cite{saeed2022image_overlapping,yan2022impact}. Our method outperformed MIL, without using any localisation labels during training, where the MIL system uses localisation labels for bag construction. Although without a statistical significance, our proposed model achieved  higher performance compared to a fully supervised baseline. Not using localisation labels in training may reduce the subjective human biases. Image-level labels may also be more time-efficient to obtain and could be obtained without image annotation e.g., through biopsy.

The RL model demonstrated more volatile performance (higher standard deviations) compared to baseline alternatives which is partly inherent to RL models. Integrating additional metrics into reward definition e.g., combining the proposed reward with image-wise prediction or dilute dice scores could introduce additional spatial information in more ambiguous cases, reducing volatility. Weak supervision may provide a clinically viable solution for prostate cancer diagnostics due to reduced bias propagation and explainable output. Furthermore, using biopsy-based labels for 3D images, may be a natural next step for such a weakly-supervised learning. 

We present a weakly supervised localisation system based on RL. A classification probability-based reward quantifying the likelihood of object presence within a localised region, is used to train a controller which learns localisation. The quantification of the likelihood of object presence is carried out by a pre-trained object-ness classifier which is trained only using binary labels of object presence at the image level, the only labels used in training. Evaluating our approach for the challenging prostate cancer segmentation task shows that our proposed method leads to an effective cancer localisation policy learnt using only image-level labels of cancer presence.

\section*{Acknowledgements}

This work was supported by the EPSRC grant [EP/T029404/1], Wellcome/EPSRC Centre for Interventional and Surgical Sciences [203145Z/16/Z], and the International Alliance for Cancer Early Detection, an alliance between Cancer Research UK [C28070/A30912; 73666/A31378], Canary Center at Stanford University, the University of Cambridge, OHSU Knight Cancer Institute, University College London and the University of Manchester.

\vfill
\pagebreak

\bibliographystyle{IEEEbib}
\bibliography{strings}

\end{document}